\documentclass[letterpaper]{article} 
\usepackage{aaai25}  
\usepackage{times}  
\usepackage{helvet}  
\usepackage{courier}  
\usepackage[hyphens]{url}  
\usepackage{graphicx} 
\usepackage{natbib}  
\usepackage{caption} 
\usepackage{algorithm}
\usepackage{algorithmic}
\usepackage{multirow}

\usepackage{newfloat}
\usepackage{listings}
\DeclareCaptionStyle{ruled}{labelfont=normalfont,labelsep=colon,strut=off} 
\floatstyle{ruled}
\newfloat{listing}{tb}{lst}{}
\floatname{listing}{Listing}
\title{Comparing Methods for Bias Mitigation in Graph Neural Networks}

\title{Comparing Methods for Bias Mitigation in Graph Neural Networks}
\author {
    Barbara Hoffmann\textsuperscript{\rm 1},
    Ruben Mayer\textsuperscript{\rm 1}
}
\affiliations {
    \textsuperscript{\rm 1}University of Bayreuth, Universitätsstraße 30, 95447 Bayreuth\\
    barbara.hoffmann@uni-bayreuth.de, 
    ruben.mayer@uni-bayreuth.de
}

\begin{document}

\maketitle

\begin{abstract}
This paper examines the critical role of Graph Neural Networks (GNNs) in data preparation for generative artificial intelligence (GenAI) systems, with a particular focus on addressing and mitigating biases. We present a comparative analysis of three distinct methods for bias mitigation: data sparsification, feature modification, and synthetic data augmentation. Through experimental analysis using the \texttt{german credit} dataset, we evaluate these approaches using multiple fairness metrics, including statistical parity, equality of opportunity, and false positive rates. Our research demonstrates that while all methods improve fairness metrics compared to the original dataset, stratified sampling and synthetic data augmentation using GraphSAGE prove particularly effective in balancing demographic representation while maintaining model performance. The results provide practical insights for developing more equitable AI systems while maintaining model performance.


\end{abstract}

\section{Introduction}

The increasing use of generative artificial intelligence (GenAI) systems has increased the importance of data preparation
to ensure fair and unbiased results \cite{sengar2024generativeAI}. Graph Neural Networks (GNNs), with their ability to process and learn from complex relational data structures, can be used as powerful tools in this preparatory phase. The effectiveness of GNNs, though, is inherently limited by the quality and representativeness of the underlying data they process \cite{min2022graph}.

Data bias presents a particular challenge in this context, as it can be amplified through the various stages of artificial intelligence (AI) system development, from initial data preparation to final model outputs. Datasets often do not reflect a perfect world scenario, underrepresenting certain groups or individuals. Such imbalances can lead to discriminatory outcomes in AI applications, if not properly addressed.
Our research investigates different distinct approaches to mitigating biases while maintaining the utility of GNNs in data preparation. \\

This paper makes the following key contributions to the field: 
\begin{itemize}
    \item A comparative analysis of different bias mitigation strategies
    \item An evaluation of their impact on both fairness metrics and model performance
    \item A balanced approach to synthetic data generation using GraphSAGE \cite{hamilton2017graphsage} that preserves network structure while improving demographic representation
\end{itemize}

\section{The Role of GNNs in Generative AI}

GNNs have the potential to significantly contribute to the preparation of data for GenAI systems. GNNs are a type of neural network specifically designed to process graph-structured data \cite{scarselli2008graph} \cite{gori2005new}, where they model both the entities as nodes and their interconnections as edges. By learning representations of these graphs or their components, GNNs are able to make predictions based on their structure \cite{vatter2023evolution}. While GNNs are not inherently generative models, they can be integrated into generative workflows in several ways.
\\ 

\begin{enumerate}
    \item \noindent \textbf{GNNs as a tool for data preparation in generative AI:} 
    GNNs can play a role in preparing data for generative models, particularly when the data exhibits complex relationships between entities. GNNs can learn the underlying structure of these relationships and provide representative graph-based features as input for generative models. By extracting meaningful information from graph data, GNNs help generative models produce more realistic and contextually appropriate results. This means the GNNs process existing graph data to create input features for generative models. The role is to understand and represent relationships, whereas the actual generation is done in the next step by a separate model.

    \item \noindent \textbf{Graph-based generative models:} 
    Generative models that utilize graph structures can also be applied in data preparation. For example, a graph-based generative model can be used to impute missing data or infer new connections in an incomplete dataset \cite{Zikas2023}, making it more suitable for subsequent machine learning tasks. The focus lies on improving existing datasets with a specific purpose for the generated content. \\
    Some generative models are built on GNNs to create new structures, such as molecular compounds or networks \cite{xia2019graph} \cite{ingraham2019generative}. The focus is on creating entirely new structure and the end goal is the generated content itself.
\end{enumerate}


\section{The Importance of Data Quality for Bias Mitigation}

Data quality is essential for the effective performance of machine learning (ML) models, as low-quality data can lead to unreliable and inaccurate predictions \cite{kariluoto2021quality}. There are different dimensions that influence the quality of data, for example:

\begin{enumerate}
    \item Completeness: The degree to which a dataset has missing values across its features \cite{budach2022effects}.
    \item Feature Accuracy: The degree to which the feature values in the dataset accurately represent the true/ground truth values \cite{budach2022effects}.
\end{enumerate}

\noindent Completeness is one of the most influential dimensions, as missing values can significantly impair performance, particularly when the training data is largely complete while the test data is not \cite{daraio2022accounting}. Feature accuracy also plays a crucial role, as errors in feature values can lead to unreliable patterns, causing the model to learn spurious relationships that reduce generalization to new data \cite{budach2022effects}. \\
\noindent Bias can emerge through these data quality dimensions in several ways. Completeness is particularly susceptible to bias, as missing data often isn't random but systematically related to underlying social, economic, or demographic factors. For instance, in financial datasets, certain populations might be underrepresented due to limited access to banking and credit services or through facing systemic barriers like gender-based income disparities. This incompleteness can create a feedback loop where models trained on such data further marginalize these groups by making less accurate predictions for them. Feature accuracy bias arises when measurement errors or data collection processes disproportionately misrepresent characteristics of specific groups. When expanding or generating data for a dataset, it is crucial to account for feature accuracy to ensure fairness and reliability. \\

\section{Methods to Mitigate Bias in GNNs} 

Graph data is frequently derived from models of the real world, but these graphs often present a skewed or incomplete picture of reality. This issue becomes particularly concerning when a dataset includes sensitive information that could lead to biased treatment of certain groups due to their underrepresentation. To establish the fairest foundation, we assume it is ideal for a dataset to contain an equal distribution of nodes with regard to the sensitive attribute. Since simply omitting a sensitive attribute, such as removing the sensitive feature, does not always suit the needs of the use case, a better approach is to adjust the dataset to ensure the feature is evenly represented across groups. Achieving equal distribution requires adjustments to the dataset, such as sparsification of the overrepresented group, augmentation of the underrepresented group, or changing features to shift group membership.

\subsection{Mitigation via Sparsification}
We propose using three sampling methods for handling graph sparsification with a focus on fairness: random sampling, stratified sampling \cite{cochran1977sampling} \cite{he2009learning}, and class weighted sampling \cite{chawla2002smote} \cite{batista2004study}. The advantage of these methods lies in their simplicity and ability to balance the dataset while retaining key information. \\
\subsection{Mitigation via Changing Features}
We assume that the dataset consists of a total of \( X \) nodes, with a certain number of nodes having feature \( O \) and a certain number having feature \( U \), where \( O \) represents the overrepresented group and \( U \) the underrepresented group. The target count for each group is \( X/2 \), and \( NC \) represents the number of nodes that need to be converted from \( O \) to \( U \). The formula used to calculate \( NC \) is:

\[
NC = O - \left( \frac{X}{2} \right)
\]

\noindent To simultaneously balance the initial attribute and another, such as \( G \) for good and \( B \) for bad classifications, the formula below determines the nodes to convert (\( NC2 \)):

\[
NC2 = |G - \left( \frac{X}{4} \right)| + |B - \left( \frac{X}{4} \right)|
\]

\subsection{Mitigation via Augmentation}
For augmentation, we assume that filling the group of underrepresented up to the number of overrepresented creates more fairness. To calculate the number to be replenished the following formula can be used:

\[
NA = O - U
\]

\section{Experimental Setup}

\begin{figure*}[h!]
    \centering
    \includegraphics[width=\textwidth]{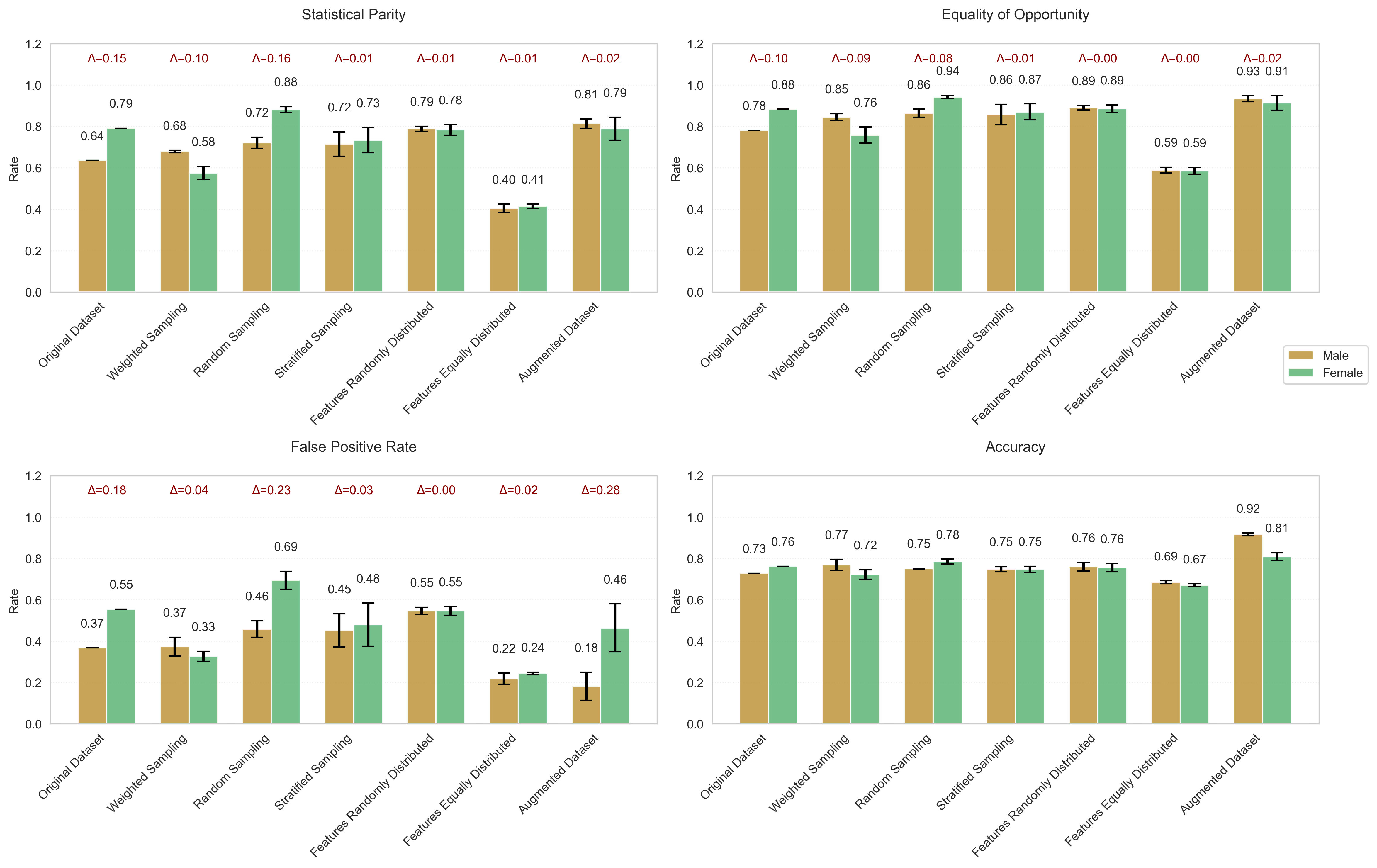}
    \caption{Fairness metric results for each modified dataset. Each method's distribution is repeated three times to account for random variation. Black bars indicate the standard deviation range, while red delta values highlight group differences. For fairness metrics, deltas are critical, whereas for accuracy, the focus is on absolute values.}
    \label{fig:fairness-metrics}
\end{figure*}

Assuming that graph data often exhibits unequal distribution due to different factors, we explore three methods to achieve a more balanced graph dataset: sparsifying the dataset, modifying its relevant features, and augmenting it with synthetic data. The resulting adjusted dataset is then each used to train a 3-layer Graph Convolutional Network (GCN) \cite{kipf2016semi} with 200 epochs, whose outputs are evaluated based on various fairness metrics. Figure \ref{fig:fairness-metrics} shows a summary of the fairness metrics and training outcomes, while Table \ref{tab:overviewDistributions} displays an overview of the gender distribution along with the distribution of the good/bad customer attribute. For all experiments, the dataset \texttt{german credit} was used, which is a commonly used machine learning dataset with 1000 nodes. It contains information about loan applications, including features like credit history, loan purpose, employment status, and personal information like gender and age, with each applicant labeled as either good or bad customer. The dataset contains data for two genders, men and women, so that we confine our study to establishing fairness between these two groups. 

\begin{table*}[t]
\centering
\caption{Overview of the distributions achieved through various sampling methods, feature modification, and dataset augmentation, applied to the german credit dataset.} 
\label{tab:overviewDistributions}
\begin{tabular}{c|c|c|c|c|c|c|c}
\hline
\multirow{2}{*} & \textbf{Original} & \textbf{Random} & \textbf{Stratified} & \textbf{Weighted} & \textbf{Adjusted Features} & \textbf{Adjusted Features} & \textbf{Augmented} \\

 & \textbf{Dataset} & \textbf{Sampling} & \textbf{Sampling} & \textbf{Sampling} & \textbf{Equally Distr.} & \textbf{Randomly Distr.} & \textbf{Dataset} \\ \hline \hline
 
\textbf{Group Sizes}   \\ \hline
male / female & 690 / 310 & 310 / 310 & 310 / 310 & 310 /310 & 500 / 500 & 500 / 500 & 690 / 690 \\ \hline

\textbf{Bad Customers}  \\ \hline
male / female & 191 / 109 & 80 / 109 & 109 / 109 & 128 / 109 & 250 / 250 & 164 / 136 & 191 / 110 \\ \hline

\textbf{Good Customers}  \\ \hline
male / female & 499 / 201 & 230 / 201 & 201 / 201 & 182 / 201 & 250 / 250 & 336 / 364 & 499 / 580 \\ \hline

\end{tabular}
\end{table*}

\subsection{Measuring Fairness} 
We selected four fairness metrics, \emph{statistical parity} \cite{dwork2012fairness}, \emph{equality of opportunity} \cite{hardt2016equality}, \emph{false positive rates}, and \emph{accuracy}, to assess algorithmic fairness from multiple perspectives. Statistical parity identifies systemic biases by analyzing the distribution of positive predictions across groups, but it does not account for predictive accuracy. Equality of opportunity addresses this by focusing on true positive rates, ensuring fair treatment. False positive rates help detect discriminatory patterns by highlighting misclassifications. Accuracy acts as a control to assess the trade-off between fairness and performance. Together, these metrics offer a comprehensive view of fairness and its impact on machine learning models. \\
When interpreting the values of the respective metric, the following guidelines must be adhered to: 
For fairness assessment the difference between the groups for parity, equality and false positive rates is the key indicator. A smaller difference means the model is more fair, regardless of the absolute values. For model performance the absolute values of accuracy and false positive rates are important. For false positive rates low values are better whereas for accuracy higher values are better.

\subsection{Sparsification of the Data Set}
Random sampling, the most basic approach, simply selects nodes randomly at a specified ratio. Stratified sampling divides the data into distinct subgroups, in this case by gender, and samples from each group independently, allowing precise control over group ratios \cite{cochran1977sampling} \cite{he2009learning}. Weighted sampling addresses class imbalances by assigning higher weights to minority classes, making underrepresented groups more likely to be selected \cite{chawla2002smote} \cite{batista2004study}. \\
The Original Dataset shows notable gender disparities, with a 15\% gap in statistical parity, a 10\% gap in equality of opportunity, and a 18\% gap in false positive rates. Random sampling maintains significant gaps with nearly the same values, 16\% gap in statistical parity, 8\% gap in equality of opportunity and 23\% in false positive rates. Weighted sampling reduces disparities, leading to a 10\% gap in statistical parity, a 9\% gap in equality of opportunity and a 4\% difference in false positive rates. Stratified sampling emerges as the most balanced approach, with minimal gender differences across all metrics - notably achieving just a 1\% gap in statistical parity and equality of opportunity, and 3\% difference in false positive rates. The accuracy of the model stays nearly the same for all three methods compared to the original dataset. \\
The measurements show that even simple sampling strategies can be used as tools for improving fairness in terms of reducing disparities between groups. While all methods succeeded in reducing disparities between groups, with the exception of random sampling for false positive rates, stratified sampling emerged as the most effective approach.

\subsection{Changing Features in the Data Set}
We apply two strategies to adjust key features in the dataset: first, we randomly reassign the gender of some male instances to female, creating a balanced 50/50 group distribution. Second, we adjust both gender and customer status, redistributing the good/bad customer labels to achieve an equal representation across groups. In our case, this method is suitable because other features do not reveal the original assigned attribute, namely male/female or good/bad customer. However it is not generally suitable as in some datasets, indirect patterns in other features may allow sensitive attributes to be inferred. \\
Both redistribution methods yield significantly improved fairness across all metrics, with differences consistently between 0\% and 2\%. The model performance slightly declines when features are equally distributed, as accuracy decreases by approximately 7\%. \\
The experiments show a trade-off between maintaining natural data patterns and achieving optimal fairness metrics. While the random and equal redistribution of features offer improvements in fairness, it serves more as a theoretical baseline for maximum achievable fairness, albeit with potential limitations in real-world applicability.

\subsection{Augmentation of the Data Set}
To augment the \texttt{german} dataset with synthetic data, we use a machine learning model capable of generating realistic profiles of additional female customers. Our approach builds a GraphSAGE neural network that learns individual customer characteristics as well as the relationships between them. We chose GraphSAGE over simpler methods like SMOTE \cite{chawla2002smote} or random oversampling because this approach preserves the network structure and relational information between data points. The network consists of two main components: an encoder that compresses customer data into a compact form and a decoder that reconstructs customer profiles from this condensed information. The model is then trained on the existing  \texttt{german credit} dataset. \\
To create new female profiles, we use a Gaussian Mixture Model (GMM) to generate data patterns similar to those of actual female customers. The GMM allows to generate synthetic data that better matches the true underlying distribution \cite{reynolds2009gaussian}. Realistic values are ensured through several checks, such as verifying loan amounts and age distributions, maintaining balanced relationships between loan amounts and approval likelihood, using ranges derived from the original dataset for these validations. \\
The results show that augmenting the dataset reduces the gap significantly to 2\% in statistical parity and equality and opportunity. For the false positive rates, on the other hand, the difference increases to 28\%. The overall model performances increases with lower absolute values in false positive rates and higher values around 80\% to 90\% in accuracy. \\
Overall, model accuracy improved without sacrificing performance, indicating successful optimization of fairness and effectiveness. Despite some remaining differences in false positive rates, synthetic data augmentation substantially mitigates bias, providing a fairer and high-performing model while maintaining a real-world structure.

\section{Related Work}



Research has addressed bias mitigation in GNNs \cite{subramonian2024theoretical, dong2022edits}, synthetic graph data generation \cite{lu2023machine, lim2016survey}, and bias prevention using synthetic data without focusing on sensitive attributes \cite{van2021decaf, jaipuria2020deflating, paproki2024synthetic}. Our work contributes by comparing strategies for bias mitigation, including graph augmentation with the GraphSAGE approach, specifically focusing on a sensitive attribute and GNNs.

\section{Conclusion}

Our experimental analysis compared three approaches to bias mitigation in Graph Neural Networks: sparsification, feature modification, and synthetic data augmentation. While all methods improved fairness metrics, stratified sampling emerged as the most balanced sparsification approach, and synthetic data augmentation using GraphSAGE demonstrated strong potential in maintaining model performance while reducing bias. Feature modification achieved significant fairness improvements but may have practical limitations. These findings provide practical insights for developing more equitable AI systems.

\bibliography{aaai25}

\begin{thebibliography}{27}
\providecommand{\natexlab}[1]{#1}

\bibitem[{Batista, Prati, and Monard(2004)}]{batista2004study}
Batista, G.~E.; Prati, R.~C.; and Monard, M.~C. 2004.
\newblock A study of the behavior of several methods for balancing machine learning training data.
\newblock \emph{ACM SIGKDD explorations newsletter}, 6(1): 20--29.

\bibitem[{Budach et~al.(2022)Budach, Feuerpfeil, Ihde, Nathansen, Noack, Patzlaff, Naumann, and Harmouch}]{budach2022effects}
Budach, L.; Feuerpfeil, M.; Ihde, N.; Nathansen, A.; Noack, N.; Patzlaff, H.; Naumann, F.; and Harmouch, H. 2022.
\newblock The effects of data quality on machine learning performance.
\newblock \emph{arXiv preprint arXiv:2207.14529}.

\bibitem[{Chawla et~al.(2002)Chawla, Bowyer, Hall, and Kegelmeyer}]{chawla2002smote}
Chawla, N.~V.; Bowyer, K.~W.; Hall, L.~O.; and Kegelmeyer, W.~P. 2002.
\newblock SMOTE: synthetic minority over-sampling technique.
\newblock \emph{Journal of artificial intelligence research}, 16: 321--357.

\bibitem[{Cochran(1977)}]{cochran1977sampling}
Cochran, W.~G. 1977.
\newblock Sampling techniques.
\newblock \emph{Johan Wiley \& Sons Inc}.

\bibitem[{Daraio, Di~Leo, and Scannapieco(2022)}]{daraio2022accounting}
Daraio, C.; Di~Leo, S.; and Scannapieco, M. 2022.
\newblock Accounting for quality in data integration systems: a completeness-aware integration approach.
\newblock \emph{Scientometrics}, 127(3): 1465--1490.

\bibitem[{Dong et~al.(2022)Dong, Liu, Jalaian, and Li}]{dong2022edits}
Dong, Y.; Liu, N.; Jalaian, B.; and Li, J. 2022.
\newblock Edits: Modeling and mitigating data bias for graph neural networks.
\newblock In \emph{Proceedings of the ACM web conference 2022}, 1259--1269.

\bibitem[{Dwork et~al.(2012)Dwork, Hardt, Pitassi, Reingold, and Zemel}]{dwork2012fairness}
Dwork, C.; Hardt, M.; Pitassi, T.; Reingold, O.; and Zemel, R. 2012.
\newblock Fairness through awareness.
\newblock In \emph{Proceedings of the 3rd innovations in theoretical computer science conference}, 214--226.

\bibitem[{Gori, Monfardini, and Scarselli(2005)}]{gori2005new}
Gori, M.; Monfardini, G.; and Scarselli, F. 2005.
\newblock A new model for learning in graph domains.
\newblock In \emph{Proceedings. 2005 IEEE international joint conference on neural networks, 2005.}, volume~2, 729--734. IEEE.

\bibitem[{Hamilton, Ying, and Leskovec(2017)}]{hamilton2017graphsage}
Hamilton, W.; Ying, Z.; and Leskovec, J. 2017.
\newblock Inductive representation learning on large graphs.
\newblock \emph{Advances in neural information processing systems}, 30.

\bibitem[{Hardt, Price, and Srebro(2016)}]{hardt2016equality}
Hardt, M.; Price, E.; and Srebro, N. 2016.
\newblock Equality of opportunity in supervised learning.
\newblock \emph{Advances in neural information processing systems}, 29.

\bibitem[{He and Garcia(2009)}]{he2009learning}
He, H.; and Garcia, E.~A. 2009.
\newblock Learning from imbalanced data.
\newblock \emph{IEEE Transactions on knowledge and data engineering}, 21(9): 1263--1284.

\bibitem[{Ingraham et~al.(2019)Ingraham, Garg, Barzilay, and Jaakkola}]{ingraham2019generative}
Ingraham, J.; Garg, V.; Barzilay, R.; and Jaakkola, T. 2019.
\newblock Generative models for graph-based protein design.
\newblock \emph{Advances in neural information processing systems}, 32.

\bibitem[{Jaipuria et~al.(2020)Jaipuria, Zhang, Bhasin, Arafa, Chakravarty, Shrivastava, Manglani, and Murali}]{jaipuria2020deflating}
Jaipuria, N.; Zhang, X.; Bhasin, R.; Arafa, M.; Chakravarty, P.; Shrivastava, S.; Manglani, S.; and Murali, V.~N. 2020.
\newblock Deflating dataset bias using synthetic data augmentation.
\newblock In \emph{Proceedings of the IEEE/CVF Conference on Computer Vision and Pattern Recognition Workshops}, 772--773.

\bibitem[{Kariluoto et~al.(2021)Kariluoto, Kultanen, Soininen, P{\"a}rn{\"a}nen, and Abrahamsson}]{kariluoto2021quality}
Kariluoto, A.; Kultanen, J.; Soininen, J.; P{\"a}rn{\"a}nen, A.; and Abrahamsson, P. 2021.
\newblock Quality of data in machine learning.
\newblock In \emph{2021 IEEE 21st International Conference on Software Quality, Reliability and Security Companion (QRS-C)}, 216--221. IEEE.

\bibitem[{Kipf and Welling(2016)}]{kipf2016semi}
Kipf, T.~N.; and Welling, M. 2016.
\newblock Semi-supervised classification with graph convolutional networks.
\newblock \emph{arXiv preprint arXiv:1609.02907}.

\bibitem[{Lim et~al.(2016)Lim, Lee, Powers, Shankar, and Imam}]{lim2016survey}
Lim, S.-H.; Lee, S.; Powers, S.~S.; Shankar, M.; and Imam, N. 2016.
\newblock Survey of approaches to generate realistic synthetic graphs.

\bibitem[{Lu et~al.(2023)Lu, Shen, Wang, Wang, van Rechem, Fu, and Wei}]{lu2023machine}
Lu, Y.; Shen, M.; Wang, H.; Wang, X.; van Rechem, C.; Fu, T.; and Wei, W. 2023.
\newblock Machine learning for synthetic data generation: a review.
\newblock \emph{arXiv preprint arXiv:2302.04062}.

\bibitem[{Min et~al.(2022)Min, Wu, Hidayetoglu, Xiong, Song, and Hwu}]{min2022graph}
Min, S.~W.; Wu, K.; Hidayetoglu, M.; Xiong, J.; Song, X.; and Hwu, W.-m. 2022.
\newblock Graph neural network training and data tiering.
\newblock In \emph{Proceedings of the 28th ACM SIGKDD Conference on Knowledge Discovery and Data Mining}, 3555--3565.

\bibitem[{Paproki, Salvado, and Fookes(2024)}]{paproki2024synthetic}
Paproki, A.; Salvado, O.; and Fookes, C. 2024.
\newblock Synthetic Data for Deep Learning in Computer Vision \& Medical Imaging: A Means to Reduce Data Bias.
\newblock \emph{ACM Computing Surveys}.

\bibitem[{Reynolds et~al.(2009)}]{reynolds2009gaussian}
Reynolds, D.~A.; et~al. 2009.
\newblock Gaussian mixture models.
\newblock \emph{Encyclopedia of biometrics}, 741(659-663).

\bibitem[{Scarselli et~al.(2008)Scarselli, Gori, Tsoi, Hagenbuchner, and Monfardini}]{scarselli2008graph}
Scarselli, F.; Gori, M.; Tsoi, A.~C.; Hagenbuchner, M.; and Monfardini, G. 2008.
\newblock The graph neural network model.
\newblock \emph{IEEE transactions on neural networks}, 20(1): 61--80.

\bibitem[{Sengar et~al.(2024)Sengar, Hasan, Kumar, and Carroll}]{sengar2024generativeAI}
Sengar, S.~S.; Hasan, A.~B.; Kumar, S.; and Carroll, F. 2024.
\newblock Generative artificial intelligence: a systematic review and applications.
\newblock \emph{Multimedia Tools and Applications}, 1--40.

\bibitem[{Subramonian, Kang, and Sun(2024)}]{subramonian2024theoretical}
Subramonian, A.; Kang, J.; and Sun, Y. 2024.
\newblock Theoretical and Empirical Insights into the Origins of Degree Bias in Graph Neural Networks.
\newblock \emph{arXiv preprint arXiv:2404.03139}.

\bibitem[{Van~Breugel et~al.(2021)Van~Breugel, Kyono, Berrevoets, and Van~der Schaar}]{van2021decaf}
Van~Breugel, B.; Kyono, T.; Berrevoets, J.; and Van~der Schaar, M. 2021.
\newblock Decaf: Generating fair synthetic data using causally-aware generative networks.
\newblock \emph{Advances in Neural Information Processing Systems}, 34: 22221--22233.

\bibitem[{Vatter, Mayer, and Jacobsen(2023)}]{vatter2023evolution}
Vatter, J.; Mayer, R.; and Jacobsen, H.-A. 2023.
\newblock The evolution of distributed systems for graph neural networks and their origin in graph processing and deep learning: A survey.
\newblock \emph{ACM Computing Surveys}, 56(1): 1--37.

\bibitem[{Xia et~al.(2019)Xia, Hu, Wang, Zhang, and Liu}]{xia2019graph}
Xia, X.; Hu, J.; Wang, Y.; Zhang, L.; and Liu, Z. 2019.
\newblock Graph-based generative models for de Novo drug design.
\newblock \emph{Drug Discovery Today: Technologies}, 32: 45--53.

\bibitem[{Zikas et~al.(2023)Zikas, Kentros, Angelis, Protopsaltis, Kamarianakis, and Papagiannakis}]{Zikas2023}
Zikas, P.; Kentros, M.; Angelis, D.; Protopsaltis, A.; Kamarianakis, M.; and Papagiannakis, G. 2023.
\newblock UniSG: Unifying entity-component-systems, 3D \& learning scenegraphs with GNNs for generative AI.
\newblock \emph{Authorea Preprints}.

\end{thebibliography}

\end{document}